\documentclass[10pt, a4paper]{article}

\usepackage{lrec-coling2024} 
\usepackage{times}
\usepackage{latexsym}
\usepackage{graphicx}
\usepackage{amsmath}
\usepackage{float}
\usepackage{booktabs}
\usepackage{authblk} 
\title{Decoding Probing: Revealing Internal Linguistic Structures in Neural Language Models using Minimal Pairs}

\name{Linyang He\textsuperscript{1}, Peili Chen\textsuperscript{2}, Ercong Nie\textsuperscript{3}, Yuanning Li\textsuperscript{2}, Jonathan R. Brennan\textsuperscript{1}} 

\address{\textsuperscript{1}University of Michigan, \textsuperscript{2}ShanghaiTech University, \textsuperscript{3}LMU Munich\\
        \{linyangh, jobrenn\}@umich.edu\\
         \{chenpl, liyn2\}@shanghaitech.edu.cn\\
         nie@cis.lmu.de}

\abstract{
Inspired by cognitive neuroscience studies, we introduce a novel `decoding probing' method that uses minimal pairs benchmark (BLiMP) to probe internal linguistic characteristics in neural language models layer by layer.  By treating the language model as the `brain' and its representations as `neural activations', we decode grammaticality labels of minimal pairs from the intermediate layers' representations. 
This approach reveals:
1) Self-supervised language models capture abstract linguistic structures in intermediate layers that GloVe and RNN language models cannot learn. 
2) Information about syntactic grammaticality is robustly captured through the first third layers of GPT-2 and also distributed in later layers. As sentence complexity increases, more layers are required for learning grammatical capabilities. 3) Morphological and semantics/syntax interface-related features are harder to capture than syntax. 
4) For Transformer-based models, both embeddings and attentions capture grammatical features but show distinct patterns. Different attention heads exhibit similar tendencies toward various linguistic phenomena, but with varied contributions.
 \\ \newline  }

\begin{document}

\maketitleabstract
\section{Introduction}
Given the exceptional performance of self-supervised language models in a wide range of NLP tasks \cite{brown2020language, devlin2018bert, radford2019language}, their ability to capture linguistic information has piqued the interest of many cognitive neuroscientists. These models offer new insights and methodologies for studying the neural mechanisms underlying language processing.

Notably, encoding analysis in cognitive neuroscience, which leverages linguistic representations from language models to reconstruct neural signals when corresponding stimulus presented, has gained popularity as a method to identify the neural substrates of speech and language processing in the brain \cite{hale2018finding, brennan2020localizing, schrimpf2021neural,li2022dissecting,goldstein2022shared}. 
However, these studies have not clearly delineated the specific linguistic information that neural language model representations capture when encoded in the brain; conclusions thus far point mostly to the contextual aspects of language processing. Therefore, determining the exact linguistic content captured by these neural language model representations is of paramount importance.

Another line of research has delved into probing linguistic capabilities. \citet{marvin-linzen-2018-targeted} introduced the concept of targeted syntactic evaluation. They designed minimal sentence pairs where two sentences differ by only one word, yet this difference renders one sentence acceptable and the other not. Here's an example: 

(1) \textit{Simple agreement}: 
\begin{tabbing}
\hspace*{1cm}a. \=  The cats \underline{annoy} Tim. (\textit{grammatical})\\
\hspace*{1cm}b. \> *The cats \underline{annoys} Tim. (\textit{ungrammatical})
\end{tabbing}

If a language model assigns a higher probability to the acceptable sentence over the unacceptable one, it is considered to have performed correctly on this task. Building on this approach, \citet{warstadt2020blimp} introduced a comprehensive minimal pairs benchmark called BLiMP covering a broad range of grammatical phenomena.
Such an extensive benchmark offers valuable insights into the holistic linguistic comprehension of a language model and help draw more concrete conclusions about the specific kinds of grammatical features learned in a certain model. 

However, previous studies using minimal pairs only focus on the probability assigned by the entire model (i.e., the final layer), not delving deeply into the linguistic characteristics inherent within the neural language model (i.e., the intermediate layers). Conversely, many encoding analysis studies in cognitive neuroscience favor the use of intermediate layer embeddings over solely the final layer. For instance, \citet{caucheteux2021disentangling} indicates that the 2/3rd layer achieves the peak encoding correlation score, while \citet{schrimpf2021neural} reveals variability in the optimal layer across participants. Given these insights, probing the linguistic information in intermediate embeddings of language models becomes crucial.

Several studies have explored the linguistic properties encapsulated in intermediate embeddings from other perspectives. \citet{tenney2019you} examined the structural insights gleaned from contextual word representations, but their approach combined information from multiple layers rather than probing each layer individually. \citet{hewitt2019structural} and \citet{manning2020emergent} further identify the emergent syntactic structures within neural networks. However, a gap still remains in comprehensively probing specific linguistic nuances across model layers, an aspect our study aims to address using minimal pairs for a more granular analysis.

Given focus on intermediate layers, the traditional method of comparing probabilities assigned by the entire language model becomes infeasible. This is because the probabilistic output of LM is based on the final layer processed through the softmax function. 
Although one could apply a softmax to an intermediate layer, there are inherent challenges. 
These layers aim to capture linguistic nuances rather than provide direct vocabulary probability distributions. Additionally, since they aren't optimized during training for token prediction, introducing softmax might distort the true linguistic patterns they represent.
As such, a new probing method for these layers is essential.

In cognitive neuroscience, when subjects are exposed to stimuli of different conditions, if stimulus categories or labels can be correctly categorized based on features derived from neural signals, it is inferred that those neural signals carry task-relevant representational information.
\cite{haxby2001distributed, mitchell2008predicting,sudre2012tracking}. Drawing inspiration from this, we treat the neural language model as a ``brain''. By feeding minimal sentence pairs into the language model simultaneously, we can obtain the internal activations corresponding to whatever distinguishes these stimuli. We designed a simple yet effective decoding probing method using these `activations' to decode grammatical or ungrammatical labels. 

Decoding probing offers a precise lens to examine the linguistic intricacies within each layer of neural language models. By applying this decoding method along with the large minimal pairs benchmark, we can pinpoint which layers excel at capturing morphology, syntax, semantics, or other linguistic phenomena. This granularity provides insights into how the model processes language hierarchically. In essence, decoding probing unlocks a deeper understanding of the inner workings of neural language models. 

In our present study, we employ decoding probing to investigate GPT-2, ELMo, and GloVe. These models exemplify self-supervised, RNN-based, and word embedding language models, respectively. Our exploration yields four principle results: 1) We first confirm that these intermediate layers support grammatical decoding that is not possible with simpler RNNs and GLoVE. 2) GPT-2 XL learns syntactic grammaticality through its initial layers, with this information being encoded among later layers. As sentences grow in complexity, more layers are needed to capture grammatical information. 3) Morphology and semantics-related features are captured by later layers than purely syntactic ones in LMs. 4) For Transformer-based models, while both embeddings and attentions capture grammatical attributes, they exhibit different patterns. Intriguingly, while different attention heads perform similarly toward various linguistic phenomena, a limited set of heads consistently emerge as predominant contributors.

\section{Decoding Probing}
\subsection{Decoding analysis in the brain}
Decoding analysis has emerged as a cornerstone methodology in cognitive neuroscience, enabling researchers to infer the specific representational content encoded in neural patterns \cite{kriegeskorte2006information}. This approach primarily revolves around the concept of training classifiers to predict specific stimulus conditions (e.g., a particular image or word) based solely on neural responses. When a classifier can predict a stimulus condition with accuracy significantly above chance, it suggests that the neural data contains information specific to that condition \cite{norman2006beyond}.
The foundational idea posits that if specific stimulus attributes can be consistently decoded from neural configurations, such configurations must inherently represent those attributes \cite{haynes2006decoding}.

One of the paradigmatic works leveraging decoding analysis in language and speech is \citet{mesgarani2012selective}. In this study, the authors employed ECoG to measure neural responses in subjects as they listened to speech.  Through the lens of decoding, the researchers could distinguish between neural patterns elicited by attended speech and normal speech, underscoring the perceptual distinctions between them. 
\begin{figure*}[t]
    \centering
    \includegraphics[width=.7\linewidth]{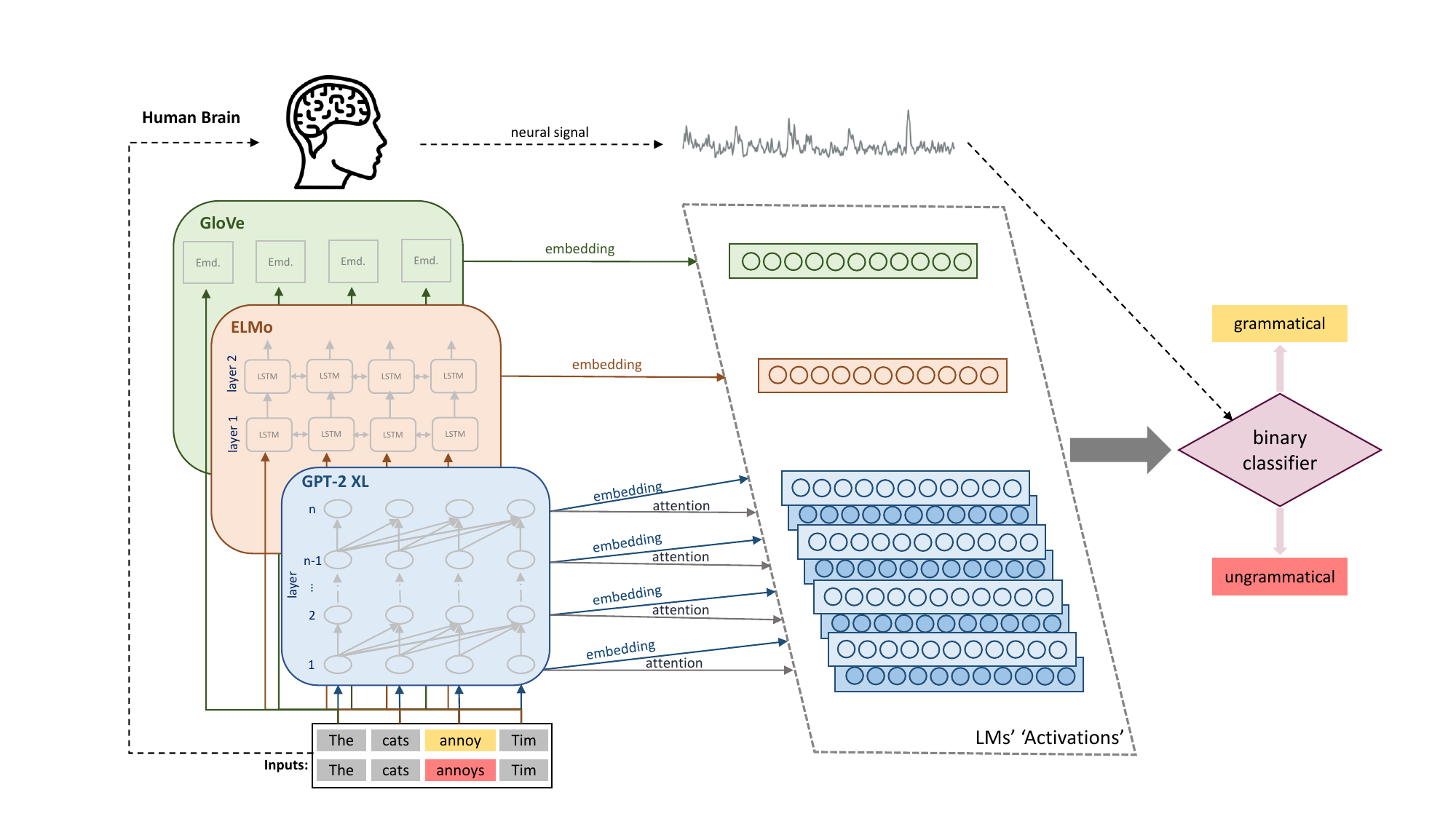}
    \caption{Decoding probing pipeline. Note: in the current study, we didn't include any actual experiments on the brain; it's just for demonstrating the core idea as the dashed lines also imply.
    }
    \label{fig:decoding}
\end{figure*}
\subsection{Decoding probing in LMs}
Building on this foundational idea and to determine the linguistic competence of each layer within the neural language models, we propose a decoding probing approach (Figure \ref{fig:decoding}).  For every minimal pair, we extracted embeddings and attentions as LMs' `activations', and we trained a binary classifier to decode whether a given sentence was grammatical or ungrammatical. This approach allowed us to investigate if specific internal regions of the model contains information related to the grammaticality of that particular minimal pair.
\section{Experimental Setup}
\subsection{Minimal pairs as model inputs}
We use minimal pairs (sentences with subtle but grammatically significant differences) as the basis for grammatical assessment. 
By comparing the assigned probabilities to grammatical and ungrammatical utterances in these pairs, we can probe if the language models truly grasp linguistic rules or merely rely on statistical biases. This method, providing a precise perspective on the linguistic comprehension of LMs, has been adopted in prior research \cite{marvin-linzen-2018-targeted, wilcox2018rnn, gauthier2020syntaxgym}.
We employed the Benchmark of Linguistic Minimal Pairs (BLiMP) dataset by \citet{warstadt2020blimp}. BLiMP is the largest and most comprehensive such minimal pairs dataset, comprising 67 tasks covering 12 phenomena.  These phenomena are divded into "morphology", "semantics-syntax interface' and "syntax" categories.  Each task presented 1, 000 sentence pairs. Within each pair one pair is grammatical and one is not. See Table \ref{tab:blimp_detail} for details. 
\subsection{Models}
\textbf{GloVe} acts as a symbolic static word embedding that encodes words into fixed vectors based on global word co-occurrence statistics regardless of specific context~\citep{pennington2014glove}.

\noindent\textbf{ELMo} is based on the LSTM architecture, which represents a type of dynamic, context-sensitive embeddings~\citep{sarzynska2021detecting}. 
For our analysis, given that ELMo's default word representation is a linear combination of two bi-LSTM layers and one char-level CNN embedding, and our intent is to investigate the performance of each individual layer, we manually extracted the LSTM activations instead of employing the standard embeddings.

\noindent\textbf{GPT-2 XL} is a self-supervised model~\citep{radford2019language}. With its Transformer architecture, GPT-2 XL is known for its ability to generate coherent text and understand complex language structures. There are 1 token-level embedding layer and 48 contextual Transformer layers in the GPT-2 XL. For each Transformer layer, there are 20 attention heads. We used both hidden states and the attention matrices to do later decoding probing. 
\subsection{Internal Activations}
\noindent \textbf{Sentence Embedding }When inputting sentences into various language models, we obtain word embeddings for each token in the sentence. The strategy for aggregating these embeddings into a coherent sentence representation varies depending on the model used. 
Given that GloVe provides static word embeddings, we employed a bag-of-words (BoW) model to construct the sentence representation. We computed the mean of the embeddings of all tokens within a sentence to generate a sentence representation. 
For ELMo and GPT-2 XL, the representation of the last token in each sentence was extracted from each layer to serve as the sentence representation for that particular layer. This decision is grounded in the understanding that the last word's embedding, especially in context-sensitive models like ELMo (a bidirectional model) and GPT-2 XL (a unidirectional model), encapsulates substantial contextual information from the entire sentence. The choice of the last word ensures that the representation has been influenced by all preceding tokens.

\noindent \textbf{Attention} We also investigate how attention matrices in GPT-2 XL captures grammatically information. For each attention head, we vectorized its attention matrix. To analyze the collective behavior, we concatenated these vectorized matrices across all attention heads to produce a final attention vector. Additionally, we examined the behavior of individual attention heads by probing their respective vectorized matrices without concatenation. 
\begin{table*}[t]
\centering
\scriptsize 
\begin{tabular}{@{}lllll@{}}
\toprule
\textbf{Level}& \textbf{Phenomenon}       & \textbf{N} & \textbf{Grammatical Example}                                                                                         & \textbf{Ungrammatical Example}                                                                                       \\ \midrule
Morphology         & Anaphor Agreement         & 2          & \textit{Many girls insulted \underline{themselves}.}                                                                 & \textit{Many girls insulted \underline{herself}.}                                                                    \\
Morphology         & Determiner Noun Agreement & 8          & \textit{Rachelle had bought that \underline{chair}.}                                                                 & \textit{Rachelle had bought that \underline{chairs}.}                                                                \\
Morphology         & Irregular Forms           & 2          & \textit{Aaron \underline{broke} the unicycle.}                                                                       & \textit{Aaron \underline{broken} the unicycle.}                                                                      \\
Morphology         & Subject Verb Agreement    & 6          & \textit{These casseroles \underline{disgust} Kayla.}                                                                 & \textit{These casseroles \underline{disgusts} Kayla.}                                                                \\
Semantics          & Quantifiers               & 4          & \textit{No boy knew \underline{fewer than} six guys.}& \textit{No boy knew \underline{at most} six guys.}                                                                   \\
Semantics/Syntax   & NPI Licensing             & 7          & \textit{The truck has \underline{clearly} tipped over.}                                                              & \textit{The truck has \underline{ever} tipped over.}                                                                 \\
Semantics/Syntax   & Binding                   & 7          & \textit{Carlos said that Lori helped \underline{him}.}                                                               & \textit{Carlos said that Lori helped \underline{himself}.}                                                           \\
Semantics/Syntax   & Control Raising           & 5          & \textit{There was \underline{bound} to be a fish escaping.}                                                          & \textit{There was \underline{unable} to be a fish escaping.}                                                         \\
Syntax             & Argument Structure        & 9          & \textit{Rose wasn’t \underline{disturbing} Mark.}                                                                    & \textit{Rose wasn’t \underline{boasting} Mark.}                                                                      \\
Syntax             & Ellipsis                  & 2          & \textit{\begin{tabular}[c]{@{}l@{}}Anne’s doctor cleans one \underline{important}  \\ \hspace{0.5cm} book and Stacey cleans a few.\end{tabular}}& \textit{\begin{tabular}[c]{@{}l@{}}Anne’s doctor cleans one book and\\ \hspace{0.5cm} Stacey cleans a few \underline{important}.\end{tabular}} \\
Syntax             & Filler Gap Dependency     & 7          & \textit{Brett knew \underline{what} many waiters find.}                                                              & \textit{Brett knew \underline{that} many waiters find.}                                                              \\
Syntax             & Island Effects            & 8          & \textit{Which \underline{bikes} is John fixing?}& \textit{Which is John fixing \underline{bikes}?}\\ \bottomrule
\end{tabular}
\caption{BLiMP dataset details. This table is adapted from \citet{warstadt2020blimp}. N denotes how many tasks are within each linguistic phenomenon. For each task, there are 1,000 pairs of sentences.}
\label{tab:blimp_detail}
\end{table*}


\subsection{Setup and Metrics}

\noindent\textbf{10-fold CV} To ensure the robustness and generalizability of our results, we adopted a 10-fold cross-validation strategy to train the logistic regression classifier. The final evaluation metric was the averaged F1 score on the held-out set across all ten validation folds.

\noindent\textbf{Feature Capture Depth} 
We explored the number of layers GPT-2 XL requires to grasp the linguistic characteristics for a particular task. To mitigate potential noise and obtain a more robust measure, we defined this layer as the first instance where the F1 score reaches 99\% of the maximum F1 score observed across all layers for that task. 


\noindent\textbf{Sentence Complexity} We measured sentence complexity, using the depth of its syntactic tree. We normalized the syntax tree depth by the length of the sentence. 
\[
\text{Sentence Complexity} = \frac{\text{Depth of Syntax Tree}}{\text{Sentence Length}}
\]
The depth of the syntax tree was computed using the SpaCy \cite{spacy2} package in Python. We then correlate sentence complexity with the depth at which GPT-2 XL effectively learns the underlying linguistic patterns.

\section{Results}
\subsection{BoW and RNN-based sentence embeddings distinguish grammaticality for some, but not all, minimal pairs}
\label{sec:3.1}
\begin{figure}[h]
    \centering
    \includegraphics[width=1\linewidth]{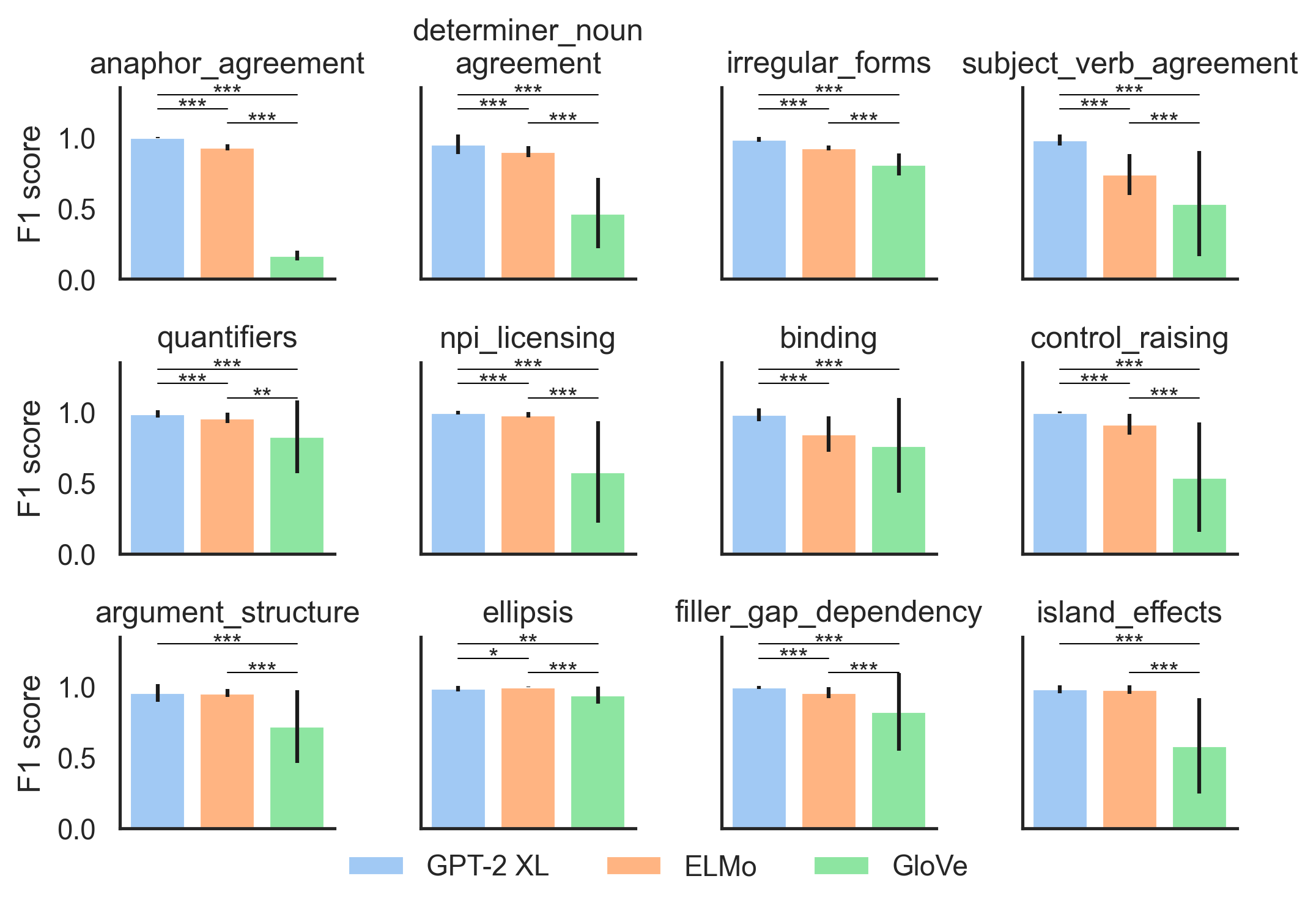}
    \caption{Decoding probing results of GPT-2 XL, ELMo and GloVe. The symbols `***', `**', and `*' denote t-test p-values less than 0.001, 0.01, and 0.05, respectively.  }
    \label{fig:gpt2_elmo_glove}
\end{figure}

For each model, we utilized the averaged F1 score across 10 cross-validation folds to denote each layer's performance. The highest score across all layers is the final F1 score for the model.  
We did a comprehensive probing analysis for GPT-2 XL, ELMo, and GloVe across all 67 linguistic tasks. A more focused result can be found in Figure \ref{fig:gpt2_elmo_glove}, which shows average performance on 12 linguistic grammatical phenomena. The top four panels represent morphology probing results, the middle four showcase the semantics-syntax interface, and the bottom four are dedicated to syntax. 
This result shows that GPT-2 XL consistently stands out, delivering superior performance across all language categories. ELMo, while lagging behind GPT-2 XL in morphology and semantics/syntax interface, shows significant closeness to GPT-2 XL for the syntatic benchmarks. Interestingly, despite its rudimentary bag-of-words approach in sentence representation, GloVe performs well in specific areas such as quantifiers and ellipses. In some tasks, GloVe is even better than ELMo. 

In the subsequent sections, we will narrow our focus to GPT-2 XL. We also excluded tasks (26 in total) where GloVe's F1 score is higher than 0.9, trying to delve deeper into the unique mechanisms by which self-supervised language models capture linguistic characteristics.

\subsection{Feature capture depth in GPT-2 XL}
\subsubsection{Grammaticality information captured through the first third of GPT2-XL layers}
\begin{figure}[h]
    \centering
    \includegraphics[width=1\linewidth]{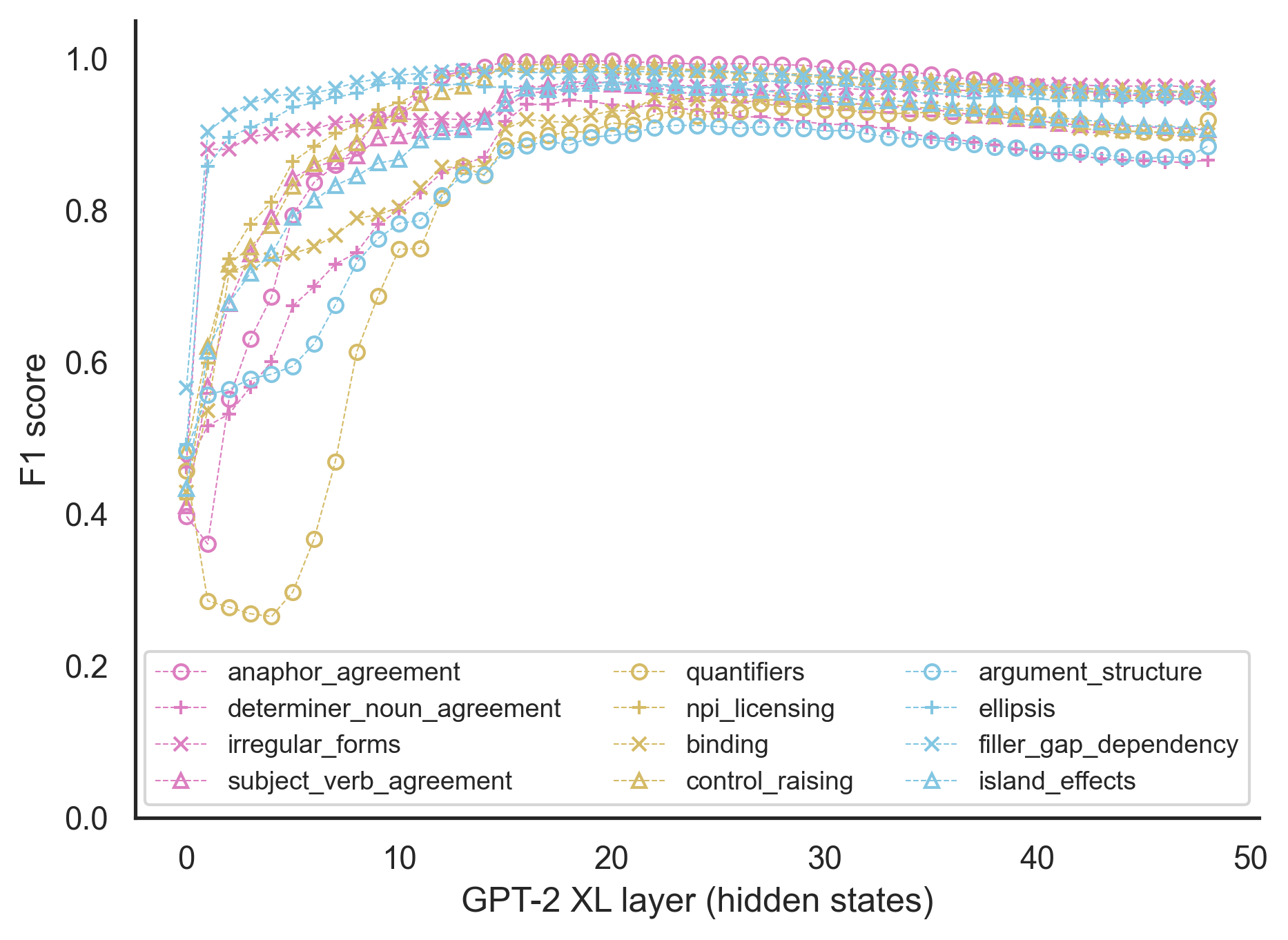}
    \caption{Decoding probing results of GPT-2 XL. Pink lines represent morphology, yellow lines highlight the semantics/syntax interface, and blue lines correspond to syntax.}
    \label{fig:blimp_gpt2}
\end{figure}
After we filtered out tasks where GloVe's F1 score is higher than 0.9, we retained tasks where GPT-2 XL's self-supervised mechanism contributes uniquely. As shown in Figure \ref{fig:blimp_gpt2}, grammatical information appears to be learned gradually through the first third of the GPT-2 XL layers. These linguistic features are also captured by subsequent layers, but with a slight downward trend.

\subsubsection{Complex sentences require more layers to capture linguistic information}

\begin{figure}[h]
    \centering
    \includegraphics[width=1\linewidth]{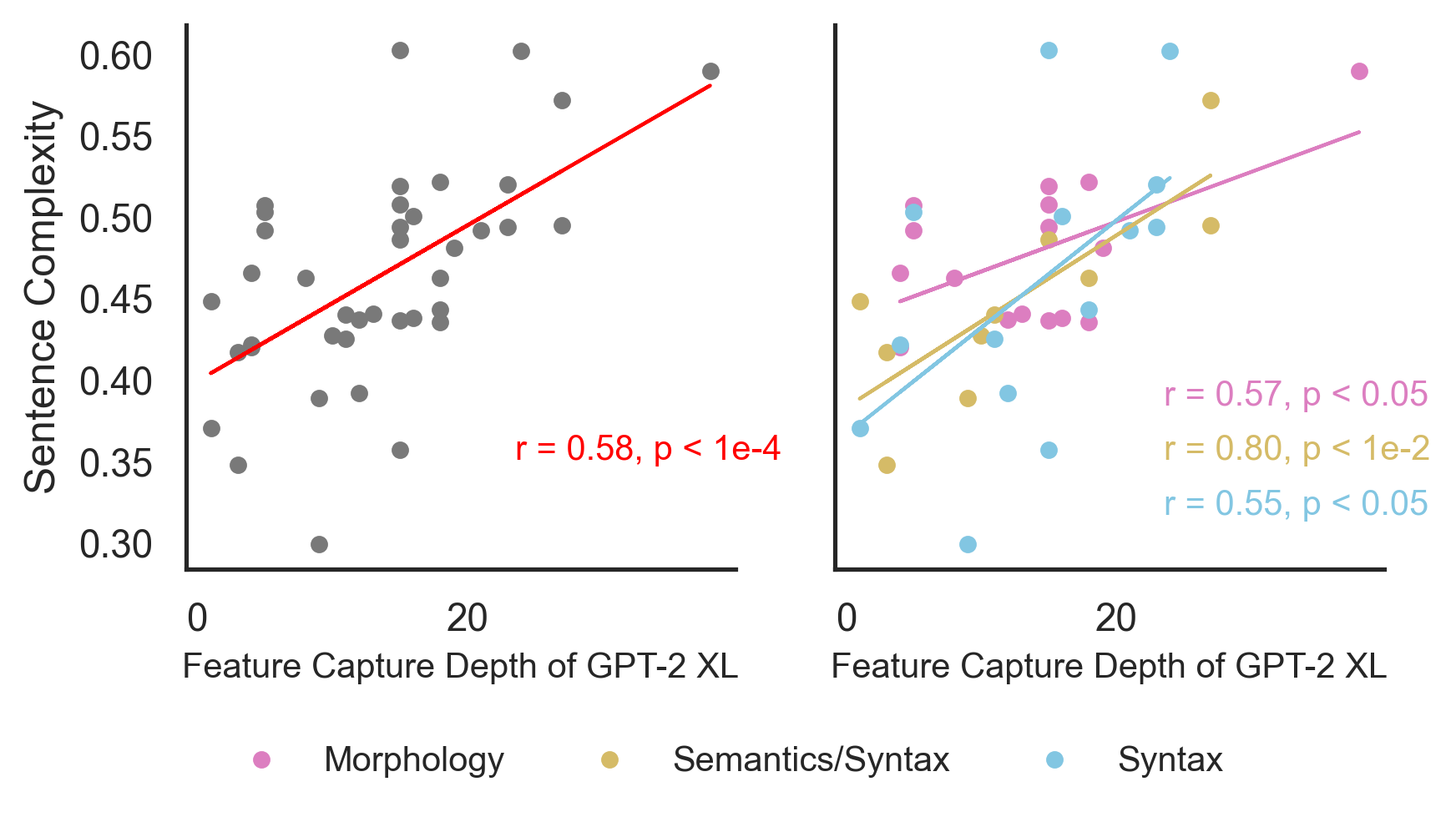}
    \caption{Sentence complexity and feature capture depth shows strong linear correlation in GPT-2 XL.}
    \label{fig:complexity}
\end{figure}
We also studied the interplay between sentence complexity and the depth required by the model to capture the corresponding linguistic information. Analyzing all the 41 tasks (after excluding 26 tasks where BoW sentence embedding model based on GloVe excelled), we compared sentence complexity against the feature capture depth for each task. The left panel of Figure \ref{fig:complexity} illustrates this relationship. A distinct linear relationship (\( r = 0.58 \) , \( p < 1 \times 10^{-4} \)) emerges between sentence complexity and the required depth for feature capture. As the complexity of the sentences increases, GPT-2 XL demands more layers to effectively internalize the associated linguistic features. 

\subsection{Semantics-syntax interface and morphology are harder to learn than syntax for LMs}

\begin{figure}[h]
    \centering
    \includegraphics[width=1\linewidth]{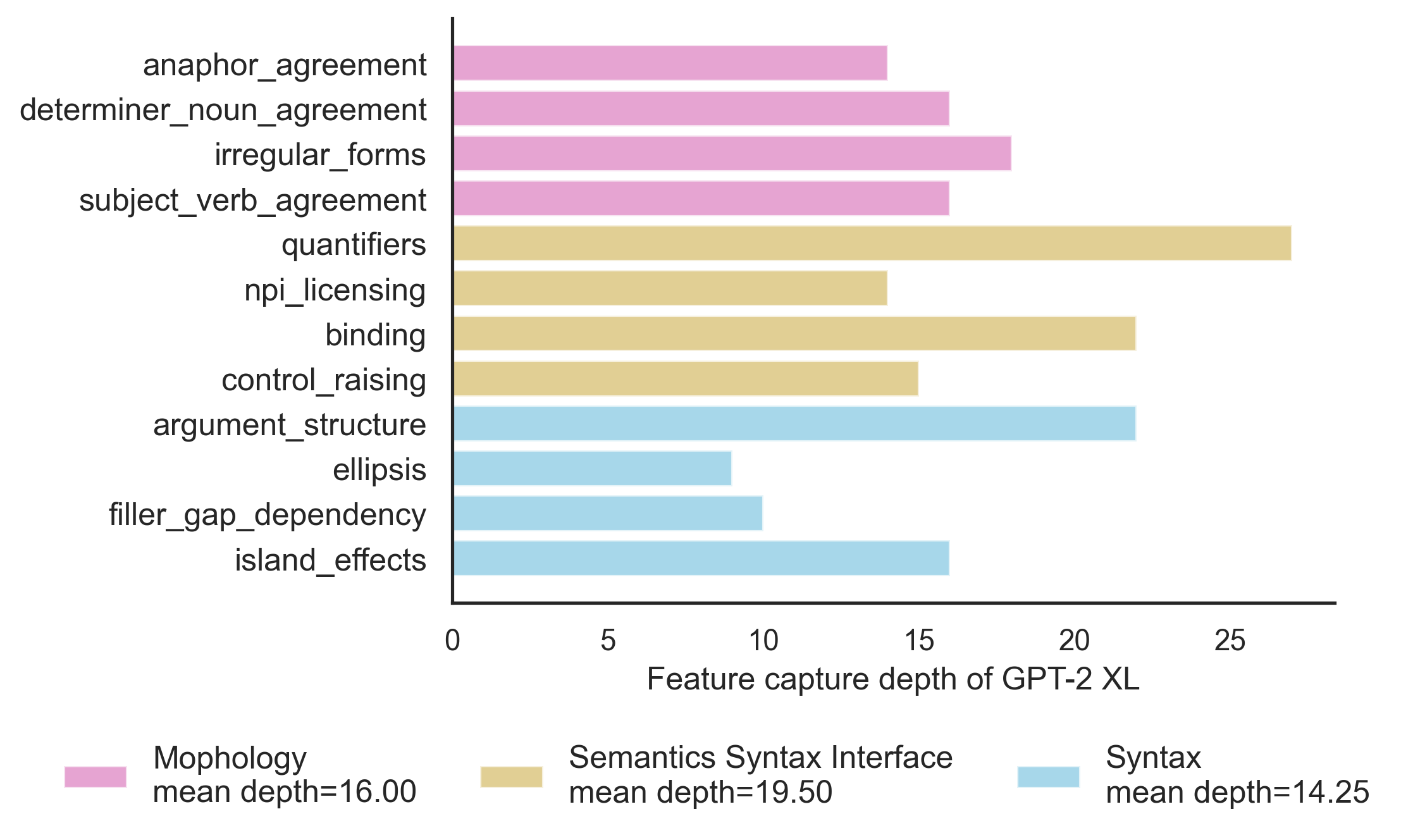}
    \caption{Feature capture depth for 12 linguistic phenomena, derived from 41 tasks after filtering out those where GloVe's F1 score exceeds 0.9. }
    \label{fig:12_levels}
\end{figure}

Figure \ref{fig:12_levels} illustrates the feature capture depth for various linguistic phenomena in GPT-2 XL. This depth is defined as the number of layers required for the model to achieve 99\% of the maximum F1 score for each phenomenon, as described in the Methods section. Here, we observe that, on average, the semantics-syntax interface demands the highest depth for efficient feature capture, suggesting its inherent complexity. Morphology follows closely, while pure syntactic phenomena can be captured through a relatively shallower depth.

\begin{figure}[h]
    \centering
    \includegraphics[width=1.0\linewidth]{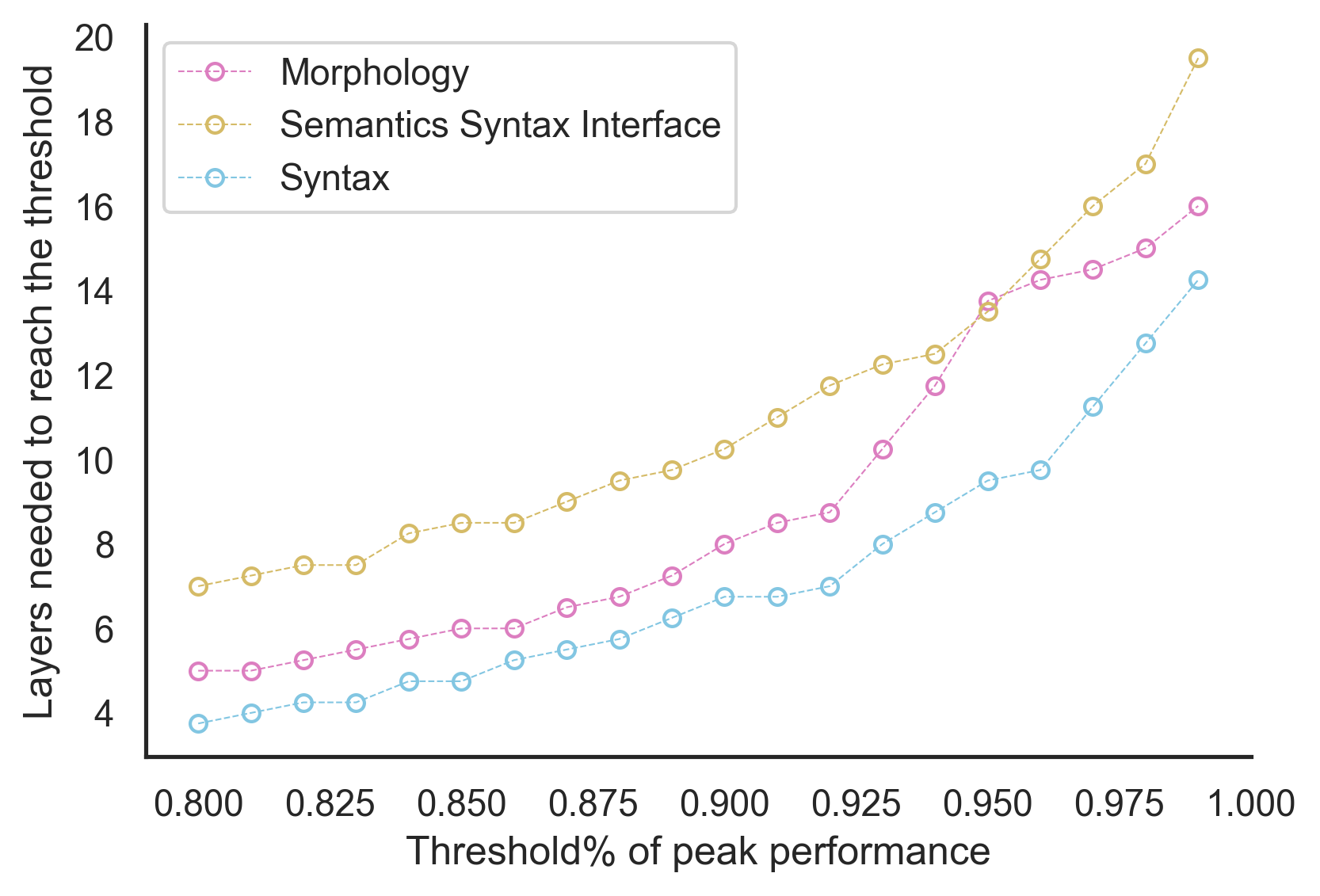}
    \caption{Average feature capture depth required by GPT-2 XL for different linguistic levels to reach certain threshold of the peak performance.  }
    \label{fig:threshold}
\end{figure}

Figure \ref{fig:threshold} further illustrates this finding. Here, the average feature capture depth required to achieve various thresholds of the maximum F1 score is plotted for the three linguistic levels. The curve corresponding to the semantics-syntax interface again consistently sits at the top, indicating that it requires more layers to reach comparable performance thresholds. Morphology occupies the middle ground, while syntax consistently requires the least depth. Roughly speaking, for GPT-2 XL to capture a comprehensive understanding of the semantics-syntax interface, it requires approximately 20 layers. For morphology, this number is around 16 layers, while syntax, being the most straightforward, needs roughly 14 layers.

To supplement these findings, the right panel of Figure \ref{fig:complexity} presents linear fits between sentence complexity and feature capture depth for each of the three linguistic levels. A clear linear relationship is evident across all three, with the semantics-syntax interface exhibiting the most pronounced correlation. Intriguingly, the slope of the linear fit offers insights into the relative difficulty of capturing linguistic features for each category. A gentler slope indicates that more layers are necessitated per unit increase in sentence complexity. Both morphology and the semantics-syntax interface showcase flatter slopes compared to syntax, reaffirming that these two linguistic levels present steeper learning challenges for GPT-2 XL. 

Results in \ref{sec:3.1}  comparing GPT-2 XL, ELMo and GloVe also suggest that features related to morphology and the semantics/syntax interface are more challenging to capture compared to syntax. 

\subsection{Attention distribution across many heads supports grammatical representations, but not much on morphology}
\begin{figure}[h]
    \centering
    \includegraphics[width=1\linewidth]{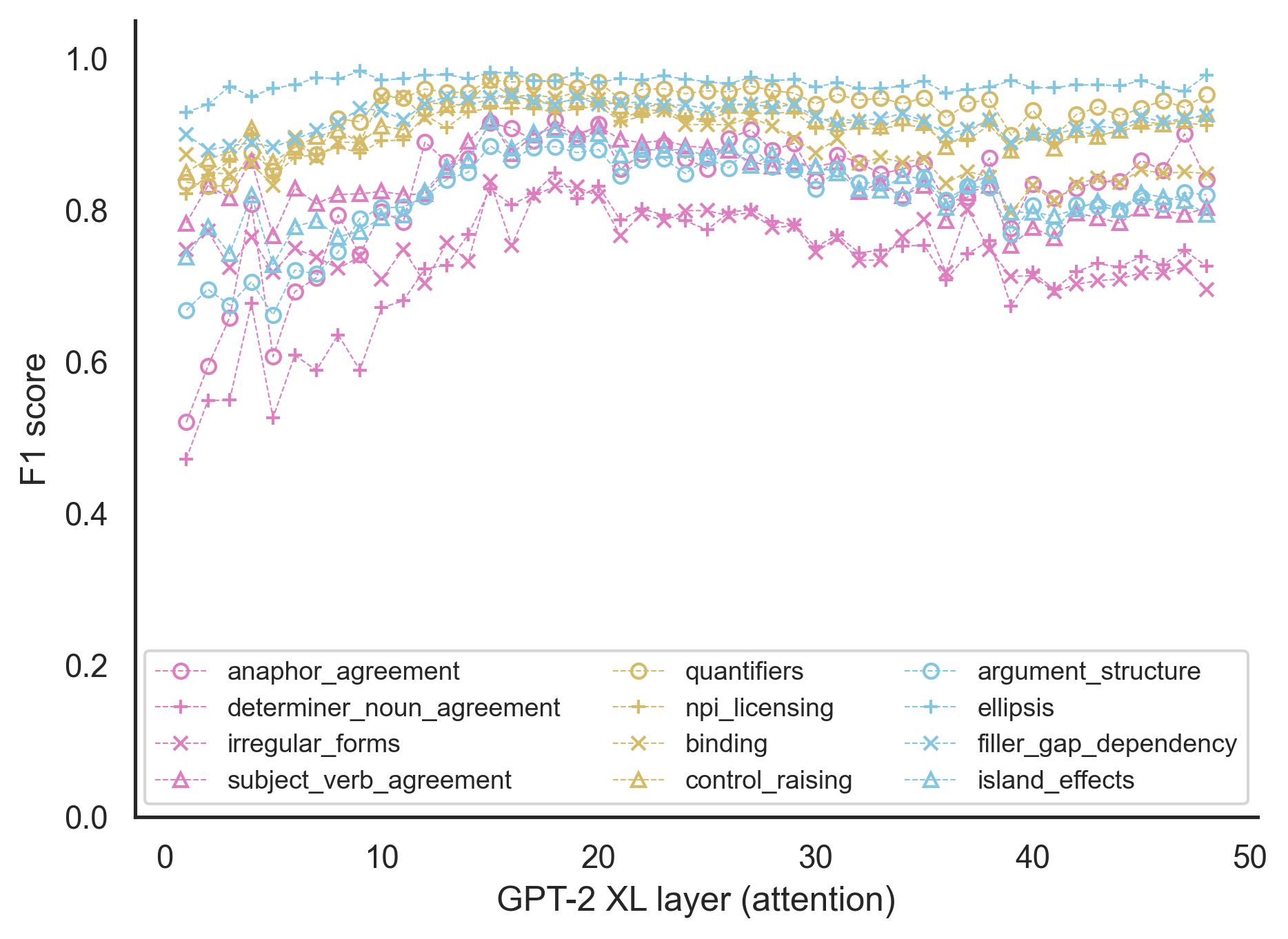}
    \caption{Decoding probing results of GPT-2 XL's attention matrices.}
    \label{fig:attention1}
\end{figure}

\begin{figure}[h]
    \centering
    \includegraphics[width=1\linewidth]{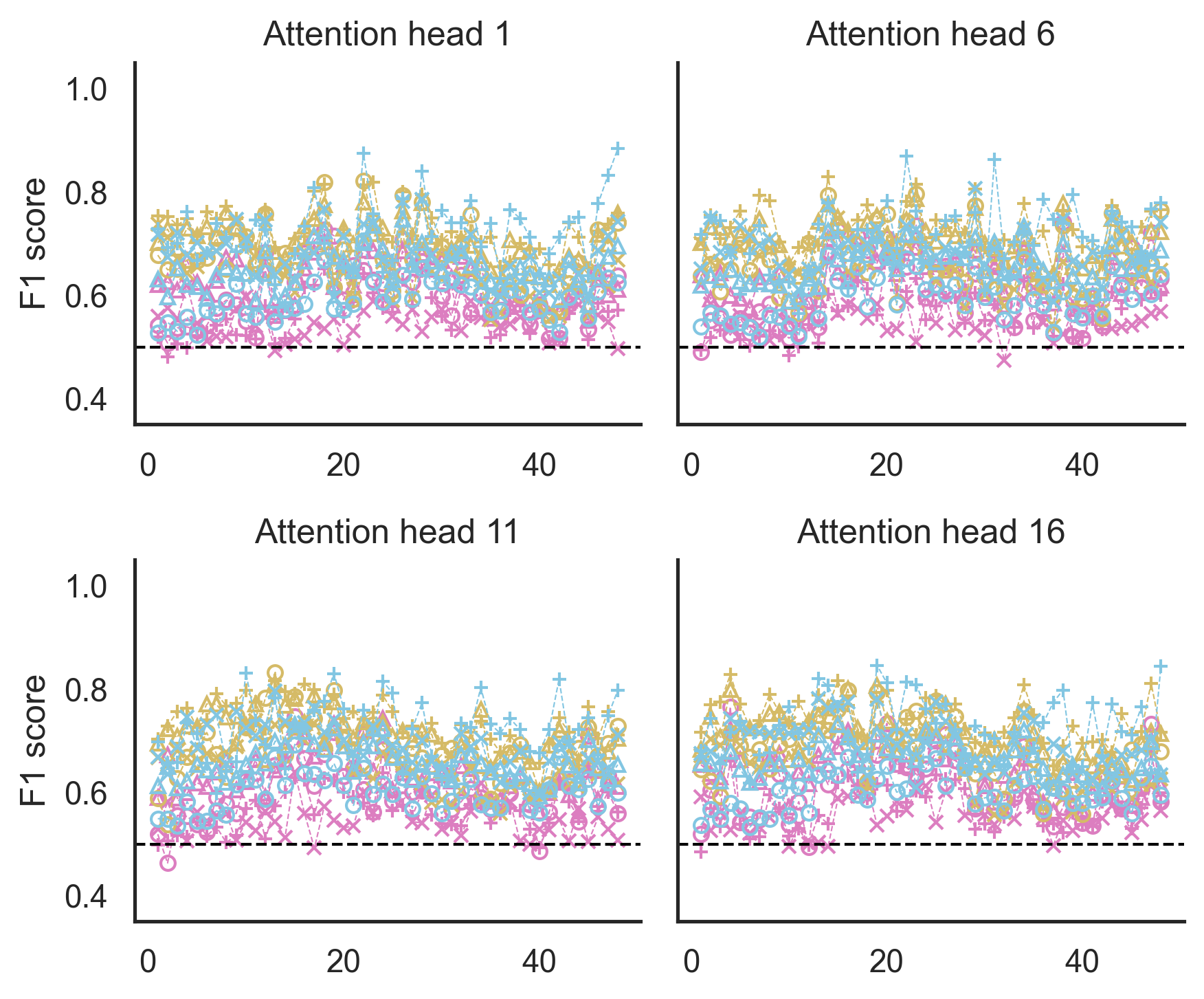}
    \caption{Decoding probing results for single attention heads. We take 4 heads as the examples. Legend is the same as in Figure \ref{fig:attention1}.}
    \label{fig:attention4}
\end{figure}

\begin{figure*}[h]
    \centering
    \includegraphics[width=0.8\linewidth]{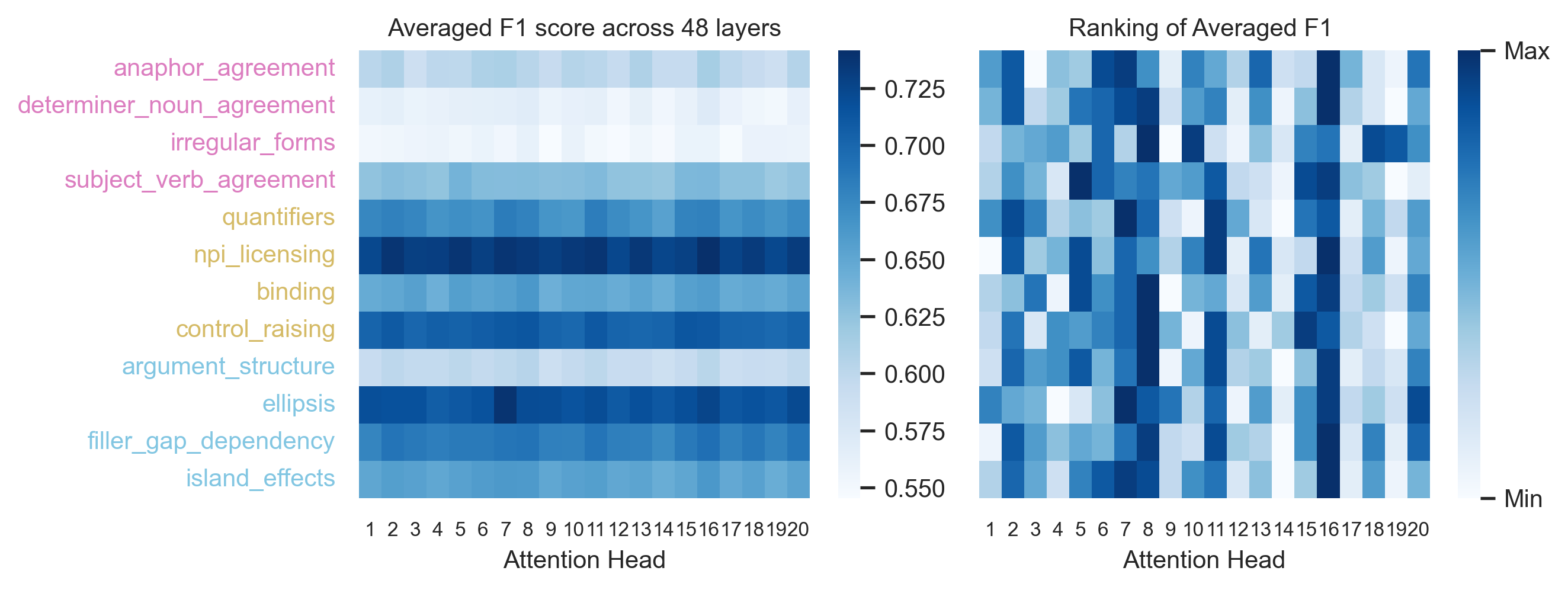}
    \caption{Performance (F1 score) and ranking of 20 attention heads across 12 linguistic phenomena. Each head is ranked based on its performance for an individual phenomenon.}
    \label{fig:attention3}
\end{figure*}

We examined the attention patterns in GPT-2 XL across its layers. Notably, attention mechanisms exhibited weaker performance in capturing morphology than in recognizing syntax or the semantics-syntax interface. Unlike the embeddings (or hidden states) which exhibit a clear gradual or incremental pattern in information capture, the concatenated attention matrices from all 20 heads did not display such a trend during decoding probing as shown in Figure \ref{fig:attention1}. This suggests that the each attention layer might operate somewhat independently, different from hidden states building upon information from previous layers. In addition, individual attention head detection strengthens this observation. The layer-by-layer performance does not show an incremental pattern, but has significant oscillations, as shown in Figure \ref{fig:attention3}.

In our exploration of individual attention heads in Figure \ref{fig:attention3}, we observed that most scores falling below 0.8. However, when the attention vectors from all heads were concatenated and examined collectively, there was a notable surge in the F1 score, exceeding 0.9. This suggests that while individual attention heads may capture specific facets of syntactic information, a comprehensive representation emerges only when the information from all heads is integrated. 

The left panel of Figure \ref{fig:attention3} shows that various attention heads yield similar performance across different linguistic phenomena. Again, their performance is notably weaker in morphology but better for syntax and the semantics-syntax interface. While the performance of individual heads appears consistent across phenomena, certain heads (e.g., No.8 and No.16, as indicated by the ranking) consistently contribute the most. However, many heads display varied contributions across different phenomena, suggesting that each head might focus on distinct linguistic facets.

\section{Discussion}
The performance between models, and across individual layers of GPT2-XL offers a granular window into the linguistic representations formed by these language models. 

\noindent\textbf{Comparison between GloVe, ELMo, GPT} The F1 scores landscape drawn from three models offers an intricate understanding of their competencies in Figure \ref{fig:gpt2_elmo_glove}. The panels delineating morphology suggest that capturing word structures might be a hurdle for ELMo, as it lags behind GPT-2 XL. Similarly, in the semantics-syntax interface, ELMo's performance indicates potential challenges in grappling with the nuances between meaning and structure. In stark contrast, its performance in syntax showcases its adeptness in capturing abstract aspects of sentence structure. GloVe's unexpected strength in specific linguistic tasks, in spite of its inability to capture structural aspects of sentences, implies that strong baselines are necessary to isolate performance aspects of more complicated models that support most abstract linguistic representations. GPT-2 XL's relatively high performance across all linguistic phenomena, from morphology to syntax, emphasizes its comprehensive linguistic capabilities. It highlights not only the versatility of the model but also its relatively strong capacity to capture certain aspects of language structure.

\noindent\textbf{GPT-2 XL's linguistic architecture} Neural networks, particularly deep ones like GPT-2 XL, are known to capture features hierarchically. In the context of language modeling, simpler linguistic features, such as basic syntax and common word relationships, are often captured in the earlier layers. More abstract and complex features, like nuanced semantics and intricate grammatical relationships, tend to be represented in the middle to later layers \cite{yosinski2015understanding}. Thus, we suggest that the gradual accumulation of grammatical information through the first third of the layers might reflect this hierarchical capture of linguistic structures in Figure \ref{fig:blimp_gpt2}.

As we move deeper into the network, the model not only captures new information but also refines and contextualizes the features from the preceding layers. This could lead to some redundancy where the same grammatical features are represented across multiple layers \cite{raghu2017svcca}. Our results support this as we see relatively high performance across all later layers in the model, albeit the slight downward trend in Figure \ref{fig:blimp_gpt2}. As the model refines and merges features, some of the explicit grammatical information captured in the earlier layers might become implicit or get overshadowed by more complex linguistic patterns. 

\noindent\textbf{Morphology and semantics harder than syntax} In essence, across models, performance thresholds, and the relationship between complexity and depth, all observations support specific hypotheses: the semantics-syntax interface and morphology present greater learning challenges for language models than pure syntax. This consensus aligns with previous studies suggesting that semantics is harder to learn than syntax \cite{tenney2019you}. Intriguingly, our results also align with neurolinguistic studies utilizing neural LMs to study the hierarchical mechanism of language processing. We speculate that the additional layers we observe for these pairs of sentences may be linked to the increased persistence of predictive representations for semantic information as reported by \citet{caucheteux2023evidence}, suggesting that semantic processing in the human brain is more long-term, high-level, than the short-term, shallow syntactic representation.

\noindent\textbf{Attention's behavior} Our findings on attention mechanisms in GPT-2 XL raise intriguing questions about their role and functionality. The absence of a clear incremental trend in attention, unlike hidden states, suggests that each attention layer in GPT-2 XL might be capturing unique linguistic information without necessarily building upon previous layers. The relatively low performance of attention heads in capturing morphology, especially given the supervised nature of our probing, might indicate that attention in GPT-2 XL isn't as adept at discerning morphological nuances. 

The variability in performance among different attention heads may suggest an inherent specialization within the model. For instance, certain heads, like No.8 and No.16, consistently outperforming others could indicate that these heads have specialized in capturing more general or prevalent linguistic features. This could be a result of the training process, where frequent patterns in the data are more likely to be captured and optimized by specific heads. 


\section{Related Work}
\noindent\textbf{Probing LM} \citet{tenney2019you} explored how much sentence structure is captured by contextually-embedded word representations. They used a series of diagnostic tasks to test whether these models can capture various grammatical and semantic features of sentences. What should be mentioned is while they did examine the outputs of LMs, their approach differed from a true per-layer probing in our study as they combined information from multiple layers so that it does not allow for a direct examination of specific linguistic information present within each layer. 

Similarly, \citet{hewitt2019structural} delved into the intricacies of neural representations by introducing a "Structural Probe". This tool specifically quantifies the degree to which word representations in a sentence capture syntactic tree distances. Hewitt and Manning's approach highlights the potential of these embeddings to encapsulate the underlying syntactic structures of sentences, suggesting that neural network architectures inherently learn syntactic relationships as a consequence of their training. \citet{manning2020emergent} further offers intriguing insights into the emergent properties of neural networks. Their study underscored that even in the absence of explicit supervision or task-specific training signals, neural architectures can spontaneously develop internal representations that echo linguistic structures.

While these studies have made significant contributions to understanding the linguistic capabilities of intermediate embeddings, they often rely on broader linguistic tests or syntactic trees, which might not offer the granularity needed to dissect the specific linguistic phenomena being captured. The use of minimal pairs provides a finer-grained lens. It allows us to pinpoint with greater precision the exact linguistic distinctions these embeddings can discern, thereby offering a more detailed understanding of the nuances in linguistic representation across different layers.

\noindent\textbf{Brain-inspired LM probing} Concurrently, cognitive neuroscientists also provide insights from neuroscience into the internal neural representation of LMs. For instance, \citet{caucheteux2021disentangling} delved into the GPT-2 embeddings and successfully isolated syntactic and semantic representations. Their findings further showed that these disentangled linguistic embeddings have cognitive neuroscience support, reinforcing the connection between neural LMs and language processing in the brain. While by using electrophysiology recording as a form of `human measurement filter', \citet{chen2023selfsupervised} suggest that intermediate layers of the deep speech model and language model share high-level contextual information.  The present effort adds specificity to the kinds of information available to the network at particular layers, which in turn contributes to understanding how those layer-wise representations may, or may not, map to human neural states.


\section{Conclusion and Future work}
Adapting the decoding theory from cognitive neuroscience to language models, as we do in our study, offers a novel approach to understanding the internal mechanisms of these models. By treating the neural language model as a "brain", we can decode its internal representations, much like how cognitive neuroscientists decode brain activity. This methodological borrowing not only bridges the gap between cognitive neuroscience and natural language processing but also paves the way for new insights into the intricacies of linguistic representation within deep learning models.

Using this method, our analysis of GPT-2 XL, ELMo, and GloVe's capabilities across various grammaticality tasks sheds light on the inherent strengths and limitations of each model. 
The hierarchically layered architecture of deep neural models like GPT-2 XL captures linguistic features progressively, with simpler ones appearing in the early layers and more complex ones in subsequent layers. Consistently, our findings emphasize that the semantics-syntax interface and morphology are more challenging for LMs compared to syntax, for both embeddings and attentions. In addition, results on individual attention heads show possible inherent specialization mechanisms.

Our current research focuses mainly on structure grammaticality, with limited attention to the conceptual capabilities of neural LMs. 
For future work, we aim to apply decoding probe methods to reveal LM's ability to master concepts using the COMPS minimal pair dataset \cite{misra2023comps}, which is designed to probe such conceptual capabilities.

\section{Bibliographical References}\label{sec:reference}
\bibliographystyle{lrec-coling2024-natbib}
\bibliography{references}

\end{document}